\documentclass[conference]{IEEEtran}
\usepackage{geometry}
\usepackage{amsmath,amsfonts}
\usepackage{algorithm}
\usepackage{array}
\usepackage[dvipsnames]{xcolor}
\usepackage{textcomp}
\usepackage{stfloats}
\usepackage{url}
\usepackage{verbatim}
\usepackage{graphicx}
\usepackage{cite}
\usepackage{bm}
\usepackage{times}
\usepackage{epsfig}
\usepackage{amssymb}
\usepackage{lipsum}
\usepackage{multirow}
\usepackage{booktabs}
\usepackage{pifont}
\usepackage{adjustbox}
\usepackage[
  bookmarks=false,
  pdfpagelabels=false,
  hyperfootnotes=false,
  hyperindex=false,
  pageanchor=false,
  citecolor=red,
  colorlinks,
]{hyperref}
\usepackage{algpseudocode}
\usepackage{bm}
\usepackage{wrapfig}
\usepackage{caption}

\begin{document}


\title{\Large\bf MINT: \underline{M}ultiplier-less \underline{INT}eger Quantization for Energy Efficient Spiking Neural Networks}

\author{\IEEEauthorblockN{Ruokai Yin,
Yuhang Li, Abhishek Moitra and
Priyadarshini Panda\\}
\IEEEauthorblockA{Department of Electrical Engineering,
Yale University. USA\\
Email: \{ruokai.yin, yuhang.li, abhishek.moitra, priya.panda\}@yale.edu}}


\maketitle

\makeatletter
\def\ps@IEEEtitlepagestyle{%
  \def\@oddfoot{\mycopyrightnotice}%
  \def\@evenfoot{}%
}
\makeatother
\def\mycopyrightnotice{%
  \begin{minipage}{\textwidth}
    \footnotesize
    ~ \hfill\\~\\
  \end{minipage}
  \gdef\mycopyrightnotice{}
}

{\small\bf Abstract---
   We propose Multiplier-less INTeger (MINT) quantization, a uniform quantization scheme that efficiently compresses weights and membrane potentials in spiking neural networks (SNNs). Unlike previous SNN quantization methods, MINT quantizes memory-intensive membrane potentials to an extremely low precision (2-bit), significantly reducing the memory footprint. MINT also shares the quantization scaling factor between weights and membrane potentials, eliminating the need for multipliers required in conventional uniform quantization. Experimental results show that our method matches the accuracy of full-precision models and other state-of-the-art SNN quantization techniques while surpassing them in memory footprint reduction and hardware cost efficiency at deployment. For example, 2-bit MINT VGG-16 achieves 90.6\% accuracy on CIFAR-10, with roughly 93.8\% reduction in memory footprint from the full-precision model and 90\% reduction in computation energy compared to vanilla uniform quantization at deployment.\footnote{Code is available at \color{magenta}{\href{https://github.com/Intelligent-Computing-Lab-Yale/MINT-Quantization}{https://github.com/Intelligent-Computing-Lab-Yale/MINT-Quantization}}}}

\section{Introduction}
Spiking Neural Networks (SNNs)~\cite{spike_nature} present a promising alternative to Artificial Neural Networks (ANNs), excelling in energy efficiency by processing sparse unary spike trains (0,1) across discrete timesteps. This efficiency makes SNNs ideal for low-power edge devices, as they operate with simplified arithmetic units. Recent backpropagation-through-time (BPTT)-based SNN efforts~\cite{direct_snn,tssl,dtsnn} have attained accuracy levels in complicated vision tasks~\cite{imagenet} that are on par with ANNs.  However, the curse of dimensionality has led to inflated model sizes for SNNs to maintain competitive accuracy, making them unsuitable for memory-constrained edge devices. While recent efforts have explored compressing BPTT-based SNNs through quantization~\cite{stq,admm,stbp_q}, specifically reducing weight memory size, two critical issues remain overlooked.

\begin{figure}[t]
\centering
\includegraphics[width=\linewidth]{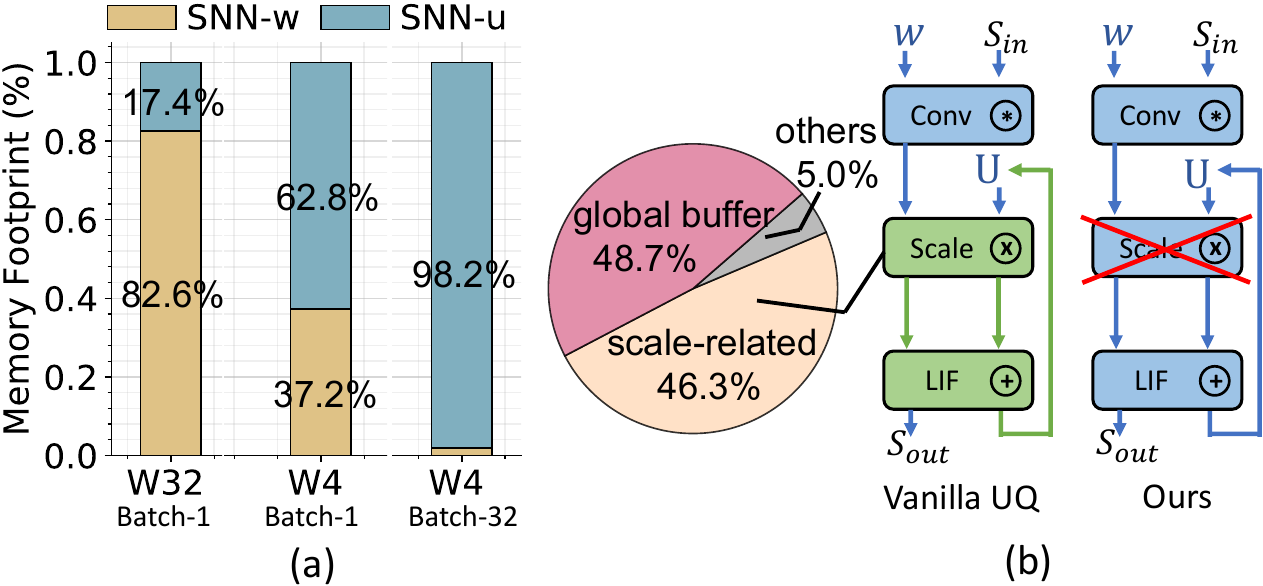}
\caption{(a): Proportion of membrane potential ($u$) in the total memory footprint of SNNs, varying with weight ($w$) precision and mini-batch size. (b): Comparison of our MINT quantization with vanilla Uniform Quantization (UQ), including a breakdown of area cost for a 4-bit vanilla UQ SNN on SpinalFlow~\cite{spinalflow}. \textcolor{Green}{Green} indicates fixed 32-bit operations, while \textcolor{CornflowerBlue}{blue} denotes quantized operations scalable with operand sizes.}
  \vspace{-5mm}
  \label{fig:intro}
\end{figure}

Firstly, prior works have not adequately addressed the size of the membrane potential in SNNs. Each Leaky-Integrate-and-Fire (LIF) neuron in an SNN requires a specific memory, known as the membrane potential, to store temporal information and generate output spikes. We observe that as weight precision decreases, membrane potentials begin to occupy a larger portion of the memory footprint, especially when using mini-batches of inputs. For instance, the proportion of the 32-bit membrane potential in the total memory footprint of the VGG-9 SNN increases from under 20\% to over 60\% as weight precision drops from 32-bit to 4-bit, as depicted on the left of Fig.~\ref{fig:intro}. Further, membrane potentials account for nearly 98\% of the overall memory footprint at a batch size of 32.

Secondly, the use of a scaling factor in Uniform Quantization (UQ)~\cite{google_quant,whitepaper_quant} by previous SNN quantization works~\cite{admm,stq} results in substantial hardware overheads during deployment. In the vanilla UQ method, the convolution result obtained by convolving quantized weights with input spikes, is multiplied by a full-precision (32-bit) scaling factor to maintain high inference accuracy, as depicted in Fig.~\ref{fig:intro}(b). This necessitates power and area-intensive multipliers in systems deploying SNNs, leading to significant hardware overheads. For example, SpinalFlow~\cite{spinalflow}, an SNN accelerator, allocates 46.3\% of its system area to support scaling factors for a 4-bit UQ SNN model, while convolution operations use only 5\% of resources, as shown in Fig.~\ref{fig:intro}. Omitting the scaling factors in UQ leads to substantial accuracy degradation at inference~\cite{whitepaper_quant}.

To address the aforementioned challenges, we introduce Multiplier-less INTeger-based (MINT) quantization, a uniform scheme for quantizing both weights and membrane potentials in SNNs. During training, MINT retains scaling factors to ensure competitive accuracy. However, during inference, we demonstrate that using a shared scaling factor for weights and membrane potentials eliminates the need for scaling factors, thereby removing 32-bit multipliers on the hardware, as illustrated in Fig.~\ref{fig:intro}(b).  Our key contributions are:

\noindent(1) We pinpoint two major hurdles in SNN quantization: the memory-intensive nature of membrane potentials and the hardware-resource demands of scaling factors. To address these, we propose MINT, which quantizes both weights and membrane potentials to extremely low precisions (2-bit) and removes the need for scaling factors during inference without sacrificing accuracy.

\noindent(2) Our experimental results show that MINT outperforms state-of-the-art methods and full-precision baselines in total memory footprint at iso-accuracy. For instance, our 2-bit quantized VGG-16 model on CIFAR-10 achieves a remarkable 93.8\% reduction in memory footprint with a negligible accuracy degradation of 0.6\%.

\noindent(3) We designed an SNN accelerator in 32nm CMOS technology to compare MINT's hardware performance with the vanilla UQ method. We also evaluated MINT on two existing SNN accelerators, SpinalFlow~\cite{spinalflow} and PTB~\cite{ptb}. Results indicate that MINT significantly conserves hardware resources, averaging an 85\% reduction in PE-array area and a 90\% reduction in computation energy compared to previous techniques, making it suitable for edge deployment.

\section{Related Work}
Quantization in SNNs has been thoroughly explored in previous studies~\cite{stq,admm,stbp_q,hessianSnn,intSnn,stdp_q,stdp_q2,qspin,conv_q1,conv_q2}, which can be classified into three groups based on the SNNs' training scheme.

\noindent \textbf{BPTT-based SNNs}\indent The ADMM method in~\cite{admm} optimizes a pre-trained full-precision network to quantize weights to low precision. Meanwhile, \cite{stq} employs K-means clustering quantization to achieve reasonable accuracy with 5-bit weight SNNs. A recent study~\cite{stbp_q} investigates fully integer-based SNN training, with fixed-point quantized weights, membrane potentials (in 8-bit to 16-bit range), and gradients. Our MINT method also belongs to this category but eliminates scaling factors and reduces both weights and membrane potentials to extremely low precision (\textit{i.e.} 2-bit).

\noindent \textbf{STDP-based SNNs}\indent Recent works on weight quantization of SNNs~\cite{intSnn,hessianSnn,stdp_q,stdp_q2,qspin} have primarily focused on shallow networks with local spike timing dependent plasticity (STDP) learning, which does not scale for complex image classification tasks. 

\noindent \textbf{Conversion-based SNNs}\indent Several works investigate quantization in the ANN-to-SNN conversion technique. Those works put their efforts into compressing the ANN activations to get better energy and accuracy performance on the converted-SNNs~\cite{conv_q1,conv_q2} while keeping the weights and membrane potentials at full or integer precision.

Additionally, prior efforts have explored pruning methods to reduce the SNN size~\cite{admm,stq}. Further, certain works have proposed reducing the number of timesteps required for processing an input which translates to energy and memory savings on hardware~\cite{dtsnn}. MINT is orthogonal and complementary to these SNN-optimization schemes. Finally, there have been separate hardware efforts to accelerate SNNs during inference with temporal/spatial reuse dataflows of weights and membrane potentials~\cite{sata,spinalflow,ptb}. MINT can be deployed on these accelerators without additional hardware resources.

\section{Preliminaries}

\begin{figure}[t]
\centering
\includegraphics[width=0.7\linewidth]{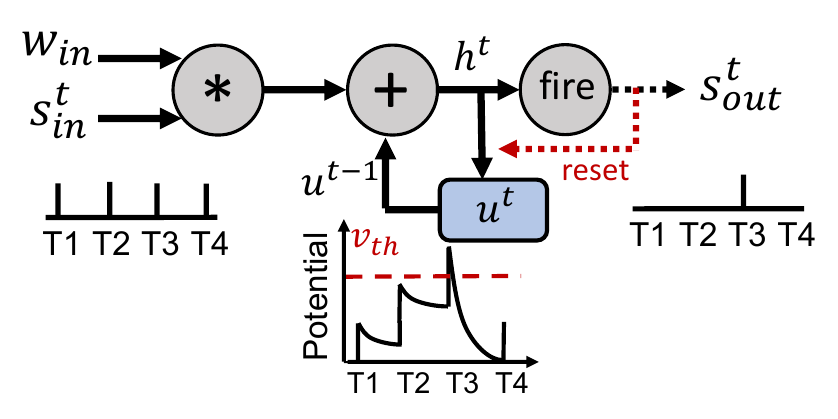}
\vspace{-2mm}
  \caption{Illustration of the behavior of the LIF neuron.}
\vspace{-4mm}
  \label{fig:intro_LIF}
\end{figure}

\noindent \textbf{Spiking LIF Neurons}\indent SNNs leverage Leaky-Integrate-and-Fire (LIF) neurons to process unary spike trains across multiple discrete timesteps. LIF neurons introduce non-linearity and capture temporal information. The behavior of LIF neurons during inference can be described in three stages. Firstly, in the `\textbf{update}' stage, the membrane potential matrix at layer $l$, denoted as $\bm{H}^{(t)}_{l}$, is updated using the weight matrix $\bm{W}_{l}$, input spike matrix $\bm{S}^{(t)}_{l-1}$ from the previous layer at timestep $t$, and the residual membrane potential matrix from the previous timestep, $\bm{U}^{(t-1)}_{l}$:
\vspace{-1mm}
\begin{equation}
\vspace{-1mm}
\label{eq:1}
\bm{H}^{(t)}_{l}=
{\:}\bm{W}_{l}\bm{S}^{(t)}_{l-1} + 
\tau\bm{U}^{(t-1)}_{l},
\end{equation}
where $\tau\in(0,1]$ is the leakage factor simulating potential decay. Next, in the `\textbf{firing}' stage, the output spike matrix $\bm{S}^{(t)}_{l}$ is determined by comparing $\bm{H}^{(t)}_l$ with a predefined threshold $v_{th}$:
\vspace{-1mm}
\begin{equation}
\vspace{-1mm}
\label{eq:2}
\bm{S}^{(t)}_{l}=
 \left\{
\begin{array}{lccl}
1 & & & {\bm{H}^{(t)}_l > v_{th}} \\
0 & & & {\textnormal{else}} .
\end{array} \right.
\end{equation}
Lastly, in the `\textbf{reset}' stage, the residual membrane potential is reset to 0 if the output spike is 1, \textit{i.e.} $\bm{U}^{(t)}_{l} = \bm{H}^{(t)}_{l}(1 - \bm{S}^{(t)}_l)$. Fig.~\ref{fig:intro_LIF} illustrates the behavior of an LIF neuron. Throughout the paper, we refer to the quantization of the membrane potential as the residual membrane potential, denoted by $\bm{U}$. For training, we use the surrogate gradient method to approximate gradients for the non-differentiable LIF neuron~\cite{tssl} and employ the cut-off approximation for BPTT training of SNNs~\cite{dtsnn}.

\noindent \textbf{Uniform Quantization}\indent Uniform integer quantization (UQ) methods, extensively studied in prior works \cite{google_quant,whitepaper_quant}, involve an affine mapping between the integer vector $\mathbf{q}$ and the floating-point vector $\mathbf{r}$, defined as follows:
\vspace{-2mm}
\begin{equation}
\vspace{-2mm}
\label{eq:4}
\mathbf{r} = \alpha \left(\mathbf{\hat{q}}-Z\right),
\end{equation}
where $\alpha$ is the scaling factor, and $Z$ is the zero-point, both of which are quantization hyperparameters. In $n$-bit quantization, $\mathbf{\hat{q}}$ contains integers representing one of the $2^n$ quantized levels in the range $[0,2^{n}]$. The scaling factor $\alpha$, a 32-bit floating-point number, scales the quantized levels to closely match the original distribution in $\mathbf{r}$. The zero-point $Z$ is an integer that ensures $\mathbf{r}=0$ is precisely represented. Interestingly, we found that omitting the zero-point does not affect the accuracy of quantized SNN models. Hence, we exclude the zero-point $Z$ in subsequent discussions.

\section{Multiplier-less Integer Quantization}
\subsection{Transform the LIF Equations}
\label{sec:3-1}
We begin by naively applying vanilla UQ to Eq.~\ref{eq:1}, using distinct full-precision scaling factors $\alpha_1$, $\alpha_2$, and $\alpha_3$ for each integer quantity. Assuming no output spike is fired at timestep $t$, the `\textbf{update}' stage equation is:
\vspace{-1mm}
\begin{equation}
\vspace{-1mm}
\label{eq:11}
\alpha _3 \hat{\bm{U}}^{(t)}_{l}=
\alpha _1\hat{\bm{W}}_l\bm{S}^{(t)}_{l-1} + 
\alpha_2 \tau\hat{\bm{U}}^{(t-1)}_{l},
\end{equation}
which can be rewritten as
\vspace{-1mm}
\begin{equation}
\vspace{-1mm}
\label{eq:12}
\hat{\bm{U}}^{(t)}_{l}=
\frac{\alpha _1}{\alpha _3}\hat{\bm{W}}_l\bm{S}^{(t)}_{l-1} + 
\frac{\alpha _2}{\alpha _3} \tau\hat{\bm{U}}^{(t-1)}_{l}.
\end{equation}

In Eq.~\ref{eq:12}, assuming $\tau=0.5$ (computable by right shift), the only non-integer multiplicands are ${\alpha _1}/{\alpha _3}$ and ${\alpha _2}/{\alpha _3}$.
In order to remove these two full-precision multiplications, we assume $\alpha_1 = \alpha_2 = \alpha_3 = \alpha$. By having all the scaling factors equal to each other, we manage to transform Eq.~\ref{eq:12} into:
\vspace{-1mm}
\begin{equation}
\vspace{-1mm}
\label{eq:13}
\hat{\bm{U}}^{(t)}_{l}=
\hat{\bm{W}}_l\bm{S}^{(t)}_{l-1} + \tau\hat{\bm{U}}^{(t-1)}_{l}.
\end{equation}

Eq.~\ref{eq:13} now consists only of integer values and operations, without requiring multiplication for the scaling factor. We again naively apply vanilla UQ to Eq.~\ref{eq:2}. We find that the `\textbf{firing}' stage generates an output spike of $1$ only if the following inequality holds true:
\vspace{-1mm}
\begin{equation}
\vspace{-1mm}
\label{eq:14}
\alpha _1\hat{\bm{W}}_l\bm{S}^{(t)}_{l-1} + 
\alpha_2 \tau\hat{\bm{U}}^{(t-1)}_{l} > v_{th}.
\end{equation}
Since $\alpha_1= \alpha_2 = \alpha$, we can divide both sides of the inequality by $\alpha$. Eq.~\ref{eq:14} is thus transformed into the form of:
\vspace{-1mm}
\begin{equation}
\vspace{-1mm}
\label{eq:15}
\hat{\bm{W}}_l\bm{S}^{(t)}_{l-1} + 
\tau\hat{\bm{U}}^{(t-1)}_{l} > \frac{v_{th}}{\alpha}.
\end{equation}

We have empirically observed that $\alpha$ is always greater than zero, which means that the direction of the inequality in Eq.~\ref{eq:15} is always preserved. Additionally, since the left-hand side of Eq.~\ref{eq:15} is always an integer, we can transform Eq.~\ref{eq:15} into the following form and still generate the same output spikes:
\vspace{-1mm}
\begin{equation}
\vspace{-1mm}
\label{eq:16}
\hat{\bm{W}}_l\bm{S}^{(t)}_{l-1} + 
\tau\hat{\bm{U}}^{(t-1)}_{l} \geq  \left\lceil \frac{v_{th}}{\alpha} \right\rceil,
\end{equation}
where $\lceil \cdot \rceil$ is the ceil operation. We define this new integer firing threshold $\lceil \frac{v_{th}}{\alpha} \rceil$ as $\theta$, a constant that can be computed offline. Empirically we find that it is enough to use less than 6 bits to represent $\theta$ for each layer.
We will use Eq.~\ref{eq:13} and Eq.~\ref{eq:16} as the new LIF equations for our MINT method during inference. 

\begin{algorithm}[h]
\caption{Inference path at $l$-th layer of the MINT-quantized SNN at timestep $t$. The leakage factor $\tau = 0.5$. }\label{alg:inference_path}
       \textbf{Input}: \\
       Input spikes {$\bm{S}_{l-1}^{(t)}$} to the layer $l$ at timestep $t$, \\
       integer weights {$\hat{\bm{W}}_{l}$} of layer $l$, \\
       integer membrane potential {$\hat{\bm{U}}_{l}^{(t-1)}$} of layer $l$ at timestep $t-1$,
       \\
       integer firing threshold $\theta$ of value $\lceil \frac{v_{th}}{\alpha} \rceil$. \\
      \textbf{Output}: \\
      Output spikes {$\bm{S}_{l}^{(t)}$} to the layer $l+1$ at timestep $t$, \\ 
      integer membrane potential {$\hat{\bm{U}}_{l}^{(t)}$} of layer $l$ at timestep $t$.
      \begin{algorithmic}[1]
        \State {${\bm{X}}^{(t)}_{l} \gets \hat{\bm{W}}_{l} \bm{S}_{l-1}^{(t)}$}
        \State {${\bm{H}}^{(t)}_{l} \gets \bm{X}^{(t)}_{l} + \bm{U}^{(t-1)}_l$\texttt{>>} 1}         
                \If{$\bm{H}^{(t)}_{l} \geq \theta$}
                    \State {$\bm{S}_{l}^{(t)} \gets 1$}
                    \State {$\hat{\bm{U}}^{(t)}_{l} \gets  0$}
                \Else
                    \State {$\bm{S}_{l}^{(t)} \gets 0$} 
                    \State {$\hat{\bm{U}}^{(t)}_{l} \gets  \bm{H}^{(t)}_{l}$}
                \EndIf
      \end{algorithmic}
\end{algorithm}
\vspace{-5mm}
\subsection{Inference Datapath}
As discussed in Sec.~\ref{sec:3-1}, sharing the scaling factor between weights and membrane potentials across timesteps for each layer allows us to implement our quantization scheme using integer-only arithmetic, eliminating the need for multiplications during inference. 

The inference algorithm is detailed in Algorithm~\ref{alg:inference_path}. First, a convolution operation is performed between the input spikes and weights. Given the unary nature of the input spikes and the quantized integer weights, this convolution operation simplifies to an integer-based accumulation. Then, the integer convolution result $\bm{X}_l^{(t)}$ is added to the quantized integer residual membrane potential $\bm{U}_l^{(t-1)}$ to compute the membrane potential $\bm{H}_l^{(t)}$. The $\bm{U}_l^{(t-1)}$ undergoes a right shift by 1, equivalent to being multiplied by a leakage factor of 0.5. Subsequently, the integer membrane potential $\bm{H}_l^{(t)}$ is compared with the pre-defined integer firing threshold $\theta$, as discussed in Sec.~\ref{sec:3-1}. The comparison determines the generation of unary output spikes. If no output spike is generated, the integer $\bm{H}_l^{(t)}$ is stored as the residual membrane potential $\bm{U}_l^{(t)}$. Conversely, if an output spike is generated, a zero is stored as the residual membrane potential.

This inference path is repeated for all other layers and timesteps in SNNs, utilizing integer arithmetic exclusively and effectively eliminating the need for multiplications. 

\subsection{Training with MINT Quantization}
\label{sec:4-3}

In the training's forward path, we apply the quantization function (refer to Eq. \ref{eq:17}) to both weights and membrane potentials. As mentioned in Sec.~\ref{sec:3-1}, our approach assumes $\alpha_1 = \alpha_2 = \alpha_3 = \alpha$. To realize this, we introduce a dummy scaling factor $\alpha$ for each layer, which can be shared between weights and membrane potentials. Consequently, the following quantization function $Q$ is applied during training:
\begin{equation}
\label{eq:17}
    Q(r,n) = \frac{\left\lfloor {\mathrm{clamp}\left(\frac{r}{\alpha},-1,1\right)}\textnormal{s}(n) \right\rceil \alpha}{\textnormal{s}(n)},
\end{equation}
\begin{align}
\label{x}
\textnormal{s}(n) &= 2^{n-1}-1, \notag \\
\mathrm{clamp}(r,a,b) &= \mathrm{min}(\mathrm{max}(r,a),b). \notag
\end{align}

Here, $r$ represents a full-precision floating-point number to be quantized, and $n$ is the number of bits assigned to the quantized integer. Since we focus on uniform quantization in this work, we use the same $n$ for both weights and membrane potentials across all layers and timesteps.

During backward propagation, we use full-precision floating-point gradients for all quantities and employ the straight-through estimator~\cite{google_quant} to approximate the derivative for the non-differentiable rounding function $\lfloor \cdot \rceil$.

Moreover, instead of assigning a static scaling factor for weights and membrane potentials as one of the hyperparameters, we make $\alpha$ a learnable parameter. For each layer, a distinct learnable full-precision scaling factor $\alpha$ is introduced. The initialization of $\alpha$ is set to $\frac{2<|w|>}{\textnormal{s}(n)}$, where $w$ denotes the initial weight values. $<\cdot>$ and $|\cdot|$ represent the arithmetic-mean and absolute value respectively. During backward propagation, we update $\alpha$ with full-precision gradients. Empirically, we observe that scaling the loss gradient of $\alpha$ by $\frac{1}{\sqrt{N_w\textnormal{s}(n)}}$ ensures rapid convergence and optimal accuracy. Here $N_w$ denotes the total number of weights in each layer.

\begin{figure}[t]
\centering
\includegraphics[width=\linewidth]{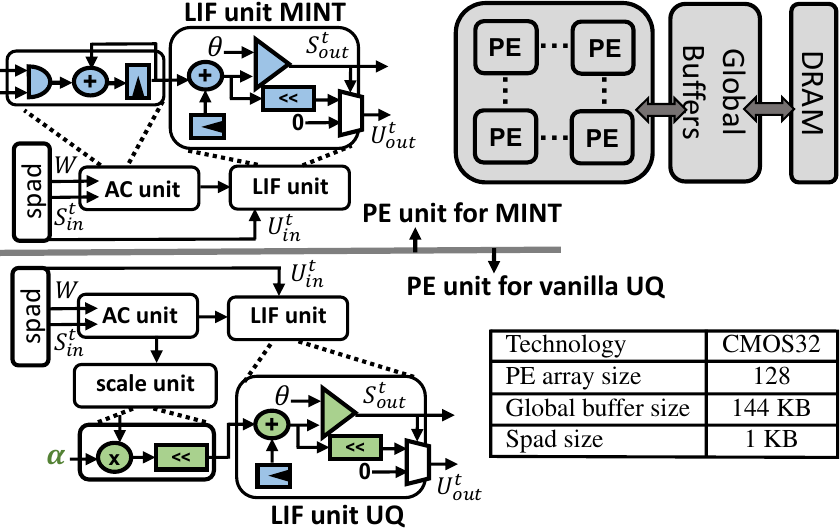}
  \caption{Systolic-array-based architecture with PE for deploying vanilla UQ models, which require scaling factors, and MINT quantized models, which do not. \textcolor{CornflowerBlue}{Blue} units are scalable with operand sizes, while \textcolor{Green}{green} units are fixed at 32-bit.}
  \label{fig:baseline_arch}
\end{figure}
\section{SNN Accelerator Design for Quantization}
\label{sec:4-1}
In recent systolic array-based SNN inference accelerators~\cite{ptb,spinalflow,sata}, low-precision weights and membrane potentials have been used to reduce memory and computation costs. However, these designs normally neglect to include the scaling units required for supporting vanilla UQ models. To facilitate a comprehensive comparison of hardware performance between MINT and vanilla UQ during inference, we design and implement a systolic-array-based SNN accelerator, as shown in Fig.~\ref{fig:baseline_arch}. The accelerator adopts the temporal output-stationary dataflow from~\cite{spinalflow,sata}, where the accumulation (AC) and LIF operations for each neuron remain stationary in a processing element (PE) across all timesteps. We adopt three levels of memory hierarchy: 1) an off-chip DRAM, 2) an SRAM-based global buffer, and 3) a register-file-based scratch pad. 

We further provide two PE designs for the two methods. For running the MINT models, each PE includes 1) an AC unit to perform the `\textbf{update}' stage, and 2) an LIF unit for the `\textbf{firing}' and `\textbf{reset}' stage. All arithmetic units in the MINT PE are colored \textcolor{CornflowerBlue}{blue}, indicating that their size can be scaled down with the precision of weights and membrane potentials. For running the vanilla UQ models, a scaling unit, comprising a 32-bit multiplier and a shifter (for approximating the 32-bit float multiplication using integer operation~\cite{google_quant}), is added between the AC and LIF units. Additionally, the LIF unit in the UQ PE requires a 32-bit operand size to accomodate the full-precision input from the scaling unit. All units denoted in \textcolor{Green}{green} have 32-bit precision. Both PE designs have two scratch pad memories for holding the weights and input spikes. A small register within the LIF unit stores the residual membrane potential. All arithmetic units in both PE designs are integer-based. We allocate 144 KB for the global buffer and 1 KB for the scratch pad memory in each PE. And we set the PE array size to 128.

\section{Experiments}
\subsection{Experimental Settings}
\noindent \textbf{Algorithm Setups.}\indent We evaluate our quantization scheme using three representative deep network architectures: VGG-9~\cite{vgg}, VGG-16~\cite{vgg}, and ResNet-19~\cite{resnet}. Our experiments involve two static visual datasets: CIFAR-10~\cite{krizhevsky2009learning} and TinyImageNet~\cite{imagenet}, and one event-based dataset, CIFAR-10 DVS~\cite{cifar10dvs}. We compare our MINT method with the full-precision baseline (\texttt{fp32}), \textit{i.e.}, the model trained with 32-bit floating-point weights and membrane potentials. The training method for MINT and the full-precision baseline is identical, except that MINT applies the quantization function from Eq.~\ref{eq:17} during forward propagation. We employ the direct encoding technique\cite{direct_snn} for training SNNs, which has proven effective in training SNNs within a few timesteps. Unless stated otherwise, all experiments use timesteps $T$ = 4 ($T=8$ for CIFAR10-DVS) and a batch size of 128. We use the Adam optimizer with a learning rate of $0.001$. All models and training codes are implemented in PyTorch, following previous work~\cite{dtsnn}.
 \noindent \\ \textbf{Hardware Setups.}\indent We synthesize the accelerator described in Sec.~\ref{sec:4-1} using Synopsys Design Compiler at 400MHz using 32nm CMOS technology. We use CACTI7.0~\cite{cacti} to model the on-chip SRAM and off-chip DRAM to obtain memory statistics. 
 Energy results are generated using SATASim, a cycle-accurate SNN energy simulator~\cite{sata}. 

 \vspace{-1mm}
\subsection{Experimental Results}
\vspace{-4mm}
\noindent \\ \textbf{Accuracy.}\indent
Table~\ref{tab:acc} summarizes the accuracy results. We evaluate MINT across three bit-width groups: 2, 4, and 8. For example, \texttt{W8U8} denotes an SNN with both weights and membrane potentials quantized to 8-bit integers. While the full-precision baseline generally outperforms in terms of accuracy across most datasets and networks, MINT remains competitive, showing less than 1\% accuracy drop in all tested scenarios. Notably, some of our \texttt{W8U8} and \texttt{W4U4} models even surpass the full-precision baseline in accuracy (e.g., a 0.2\% increase on VGG-16 TinyImageNet), which suggests MINT may serve the purpose of regularization.

\begin{figure}[h]
\centering
\includegraphics[width=\linewidth]{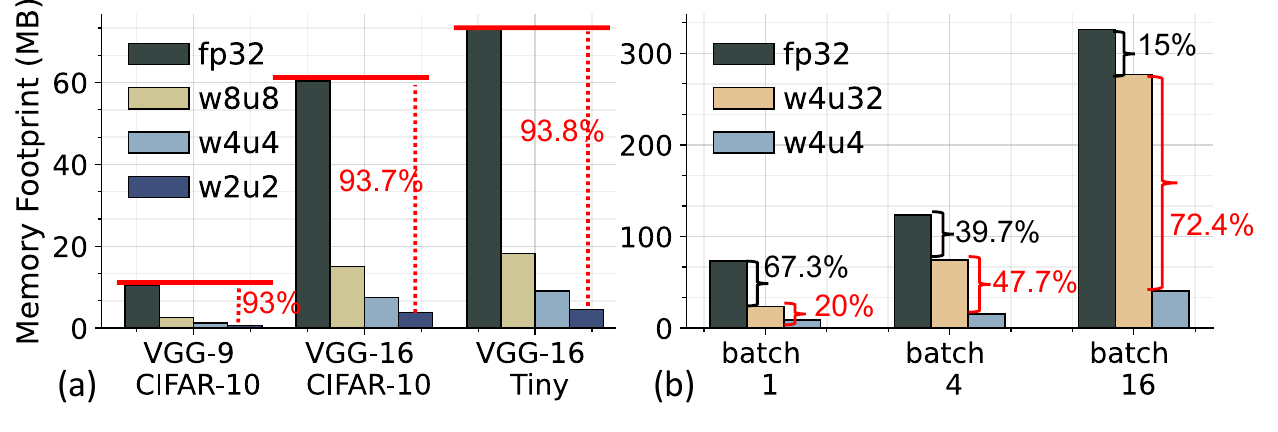}
\vspace{-6mm}
  \caption{Comparison of total memory footprint between full-precision and MINT-quantized models. (a) Overall memory reduction with a batch size of 1. (b) Memory reduction portion for different batch sizes.}
  \label{fig:memory_save}
  \vspace{-3mm}
\end{figure}

\begin{figure}[t]
\begin{center}
\begin{tabular}{@{}c@{}c}
\includegraphics[width=0.5\linewidth]{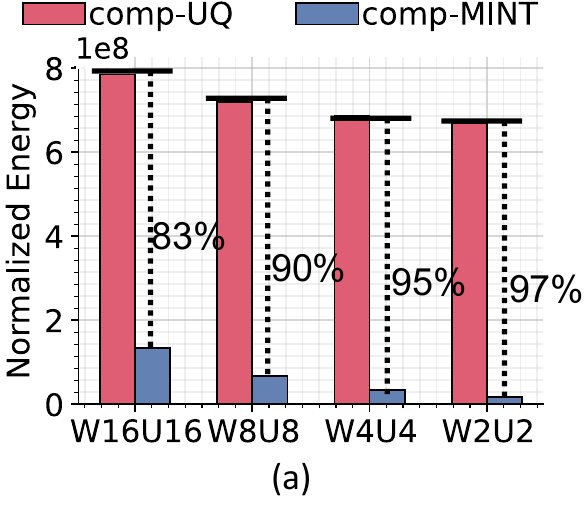} &
\includegraphics[width=0.5\linewidth]{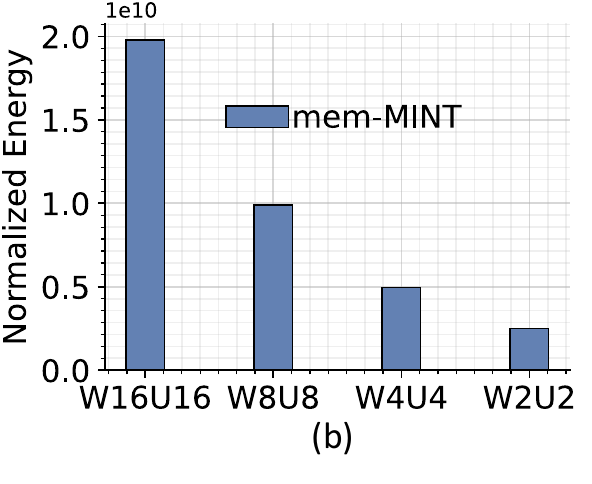} 
\end{tabular}
\end{center}
   \vspace{-7mm}
  \caption{(a) Normalized computation energy comparison between MINT and vanilla UQ on VGG-9 CIFAR-10. (b) Normalized memory energy cost of MINT for different operand precisions.}
  \label{fig:energy}
  \vspace{-4mm}
\end{figure}

\begin{table}[h]
    \centering
    \caption{Accuracy comparison between MINT and full-precision baselines across different precisions.} \label{tab:acc}
    \begin{adjustbox}{width=0.9\linewidth}
    \begin{tabular}{llccc}
    \toprule
    \multirow{2}{*}{\textbf{Method}}&\multirow{2}{*}{\textbf{\texttt{fp32}}} & \multicolumn{3}{c}{\textbf{Precision (W - U)}}\\
     & & 8-8 & 4-4 & 2-2 \\
    \midrule
     VGG-9 (CIFAR-10) & & \multicolumn{3}{c}{Top1-Accuracy (\%)}\\
     \textbf{MINT (Ours)} &88.03& 87.48  & 87.37 & 87.47\\
     \midrule
     VGG-16 (CIFAR-10)  && \multicolumn{3}{c}{Top1-Accuracy (\%)}\\
     \textbf{MINT (Ours)} & 91.15& 90.72  & 90.65 & 90.56\\
     \midrule
     ResNet-19 (CIFAR-10)  && \multicolumn{3}{c}{Top1-Accuracy (\%)}\\
     \textbf{MINT (Ours)} &91.29& 91.36  & 91.45 & 90.79\\
     \midrule
     VGG-16 (TinyImageNet)  && \multicolumn{3}{c}{Top5-Accuracy (\%)}\\
     \textbf{MINT (Ours)} &73.71& 73.92  & 73.33 & 73.18\\
     \midrule
     ResNet-19 (CIFAR10-DVS)  && \multicolumn{3}{c}{Top5-Accuracy (\%)}\\
     \textbf{MINT (Ours)} &94.2& 94.4  & 94.5 & 93.7\\
    \bottomrule
    \end{tabular}
    \end{adjustbox}
\end{table}
\noindent \\ \textbf{Memory Saving.}\indent In this section, we first present the memory footprint reduction achieved by MINT for a batch size of 1. As depicted in Fig.~\ref{fig:memory_save}(a), our \texttt{W2U2} models, on average, yield over 93\% reduction in total memory footprint. However, to enhance inference speed, previous SNN works~\cite{bntt, direct_snn, tssl} often employ mini-batch sizes greater than 1. In such scenarios, the quantization of membrane potential becomes critical. As shown in Fig.~\ref{fig:memory_save}(b), the memory overhead of the membrane potential grows significantly with larger batch sizes. For instance, for a MINT-quantized VGG-16 model on TinyImageNet with a batch size of 16, merely compressing the weight from 32-bit to 4-bit results in a 15\% reduction in total memory footprint. In contrast, further quantizing the membrane potential to 4 bits leads to an additional 72.4\% memory footprint reduction. We anticipate even more pronounced benefits from membrane potential quantization with larger batch sizes (greater than 64).
\noindent \\ \textbf{Energy Saving.}\indent We examine the difference in inference energy consumption between MINT and the vanilla UQ method using the accelerator design proposed in Sec.\ref{sec:4-1}. We normalize the results with the energy cost of a 16-bit integer multiply-accumulate operation. Fig.~\ref{fig:energy}(a) demonstrates that the computation energy of the UQ model remains relatively constant despite reductions in operand size, due to the energy-intensive full-precision multipliers needed for scaling factors. Conversely, MINT-quantized models, which do not require multipliers within PEs, exhibit a substantial decrease in computation energy as operand precision decreases. On average, MINT consumes 90\% less computation energy compared to the vanilla UQ method. Fig.~\ref{fig:energy}(b) highlights the memory energy savings achieved through quantization. For instance, our MINT-quantized \texttt{W2U2} VGG-9 network attains an approximately 87.3\% reduction in memory energy on CIFAR-10 compared to the \texttt{W16U16} model.
\noindent \\ \textbf{Comparison to Prior Works.}\indent We benchmark MINT against several state-of-the-art BPTT-based SNN quantization methods on the CIFAR-10 dataset, including STBP-Quant~\cite{stbp_q}, ST-Quant~\cite{stq}, and ADMM-Quant~\cite{admm}. We ensure consistency in weight precision, number of mini-batches, and number of timesteps ($T=8$) across all methods. Table~\ref{tab:related_acc} reveals that our method is the first to explore compressing both weights and membrane potentials to extremely low precision ($<$ 8 bits) while maintaining accuracy comparable to previous works.


\begin{table}[h]
\centering
\renewcommand*{\arraystretch}{1.2}
\caption{Accuracy and total memory footprint comparison to prior state-of-the-art SNN quantization work on CIFAR-10.}
\label{tab:related_acc}
\begin{adjustbox}{max width =0.95\linewidth}
\begin{tabular}{lcccc}
\toprule
\begin{tabular}[c]{@{}l@{}}Method\\ (CIFAR-10)\end{tabular} &
  \begin{tabular}[c]{@{}c@{}}Precision\\ (W / U)\end{tabular} &
  \begin{tabular}[c]{@{}c@{}}Accuracy (\%)\\ Top-1\end{tabular} &
  \begin{tabular}[c]{@{}c@{}}Mini\\ Batches\end{tabular} &
  \begin{tabular}[c]{@{}c@{}}Memory\\ Footprint (MB)\end{tabular}\\ \midrule
STBP-Quant                    & 8 / 14 & 86.65  & 50 & 353.79                   \\
\textbf{MINT (Ours)} & 8 / 8  &  \textbf{88.25} & 50 &\textbf{95.41}  \\ \midrule
ST-Quant                      & 5 / 32 & \textbf{88.6}  & 32 &  751.04                  \\
\textbf{MINT (Ours)} & 5 / 5  &   88.04    &   32&         \textbf{59.62}         \\ \midrule
ADMM-Quant                    & 4 / 32 &  \textbf{89.4}  &   50 & 1279.66 \\
STBP-Quant                    & 4 / 10 & 84.99 &   50&         248.39          \\
\textbf{MINT (Ours)} & 4 / 4  &   88.12    &        50&     \textbf{47.71}         \\ \midrule
ADMM-Quant                    & 2 / 32 & \textbf{89.23} &        50&    1264.85          \\
STBP-Quant                    & 2 / 8  & 33.53 &       50&     195.68          \\
\textbf{MINT (Ours)} & 2 / 2  & 88.39 &        50&    \textbf{23.85}          \\ \bottomrule
\end{tabular}
\end{adjustbox}
\end{table}

\begin{figure}[t]
\centering
\includegraphics[width=0.85\linewidth]{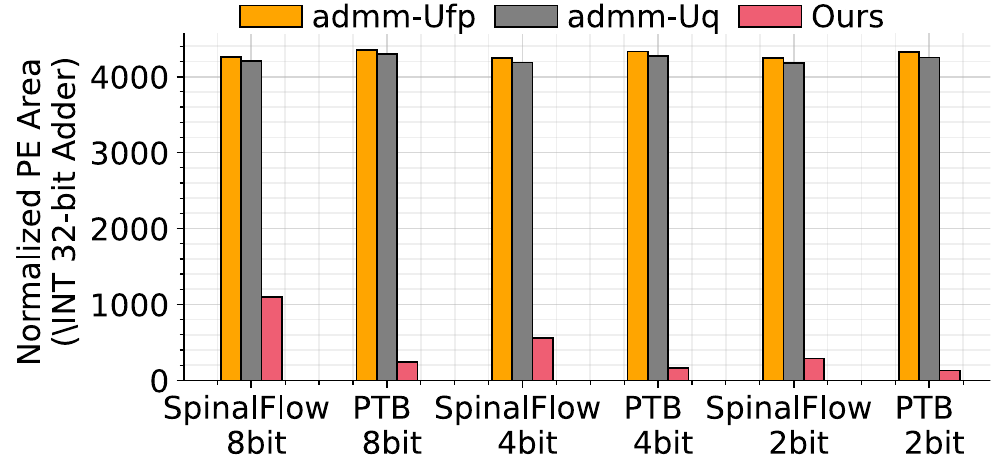}
\vspace{-1mm}
  \caption{Area comparison between MINT and prior SNN quantization methods using scaling factors.}
  \label{fig:area_cost}
  \vspace{-5mm}
\end{figure}

We also showcase that, unlike previous quantization methods that depend on scaling factors, our MINT approach is agnostic to the hardware design at deployment time and incurs no additional hardware overheads. We synthesize two existing SNN inference accelerators SpinalFlow~\cite{spinalflow} and PTB~\cite{ptb}, with two PE designs supporting both MINT and ADMM-Quant as described in Sec.~\ref{sec:4-1}. Fig.~\ref{fig:area_cost} illustrates the differences in PE-array level area. When compared to the original ADMM-Quant, which uses \texttt{fp32} membrane potentials (\texttt{admm-Ufp}), MINT achieves a reduction of 76\% (93\%), 86\% (95\%), and 93\% (96\%) in PE-array area with 8-bit, 4-bit, and 2-bit weights on SpinalFlow (PTB), respectively. We also compare ADMM-Quant with membrane potentials having the same precision as the weights (\texttt{admm-Uq}). The trend of area reduction persists, as ADMM-Quant still depends on scaling factors.

\begin{figure}[h]
\centering
\includegraphics[width=0.7\linewidth]{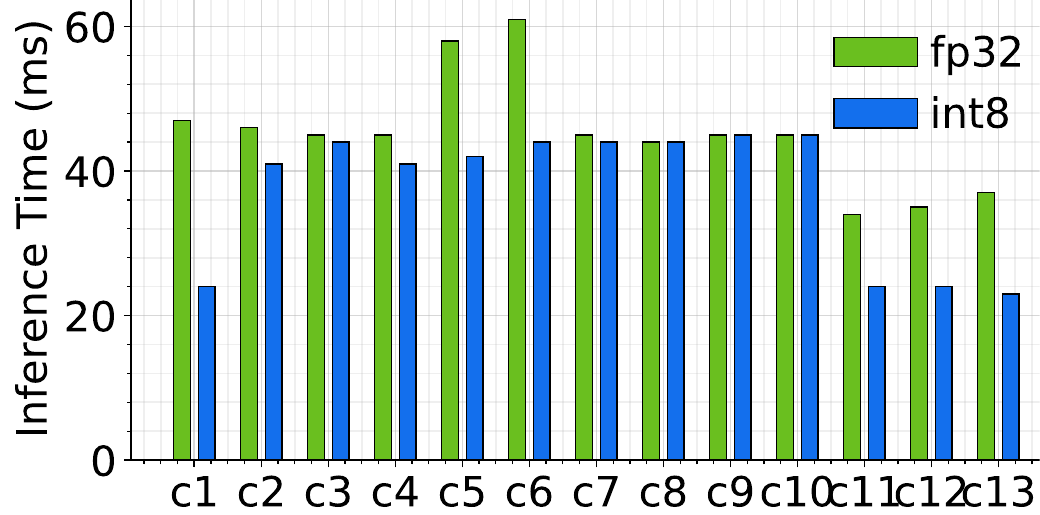}
  \caption{Layer-wise speedup of VGG-16 on TinyImageNet on NVIDIA A100 GPU.}
  \label{fig:speed_up}
  \vspace{-4mm}
\end{figure}
\subsection{Ablation Studies}
In Fig. \ref{fig:speed_up}, we compare the layerwise inference latency of our \texttt{W8U8} MINT model with the full-precision baseline for VGG-16 on TinyImageNet, implemented using CuDNN and tested on an NVIDIA A100. Notably, the \texttt{W8U8} model consistently outperforms the full-precision baseline in terms of speed on the first two layers and shows substantial acceleration on the last three layers. Overall, MINT achieves a 17.4\% acceleration for TinyImageNet on the A100.
\begin{figure}[h]
\centering
\includegraphics[width=\linewidth]{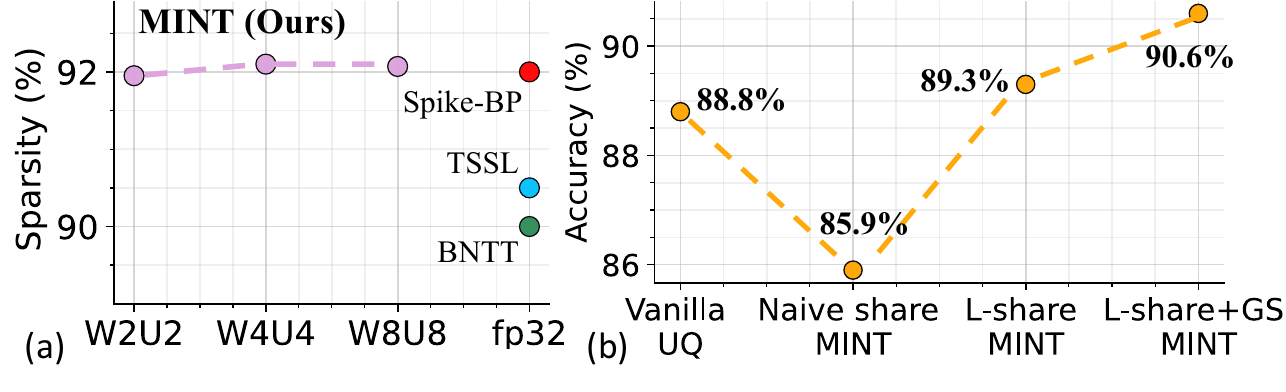}
  \caption{(a) Average spike sparsity of MINT versus other full-precision SNN works: BNTT\cite{bntt}, TSSL\cite{tssl}, and Spike-BP\cite{spike-bp}. (b) Final test accuracy comparison.}
  \vspace{-5mm}
  \label{fig:ablation_1}
\end{figure}

In Fig. \ref{fig:ablation_1}(a), we display the average spike sparsity of our MINT-quantized VGG-9 models on CIFAR-10 alongside other state-of-the-art full-precision SNN works. The results indicate that our method does not lead to any reduction in sparsity across different model precisions. By maintaining high spike sparsity comparable to other SNN works, our quantized models can benefit from similar computation energy reductions\cite{sata,bntt}. In Fig. \ref{fig:ablation_1}(b), we compare the final test accuracy of VGG-16 on CIFAR-10 between vanilla UQ and MINT with various optimization techniques applied. In the Naive share MINT, we use a predetermined scaling factor for weights and membrane potentials, resulting in a nearly 3\% accuracy drop from the vanilla UQ. Making the shared scaling factors learnable (L-share MINT) recovers the accuracy by 3.4\%. Additionally, scaling the gradient of the learnable shared scaling factors (L-share+GS), as discussed in Sec. \ref{sec:4-3}, leads to a further 1.3\% increase in accuracy.
\begin{figure}[h]
\begin{center}
\begin{tabular}{@{}c@{}c}
\includegraphics[width=0.5\linewidth]{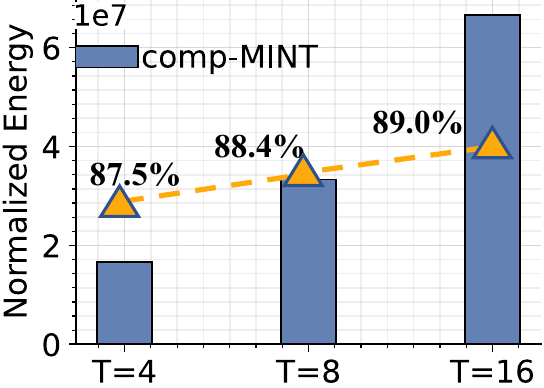} &
\includegraphics[width=0.5\linewidth]{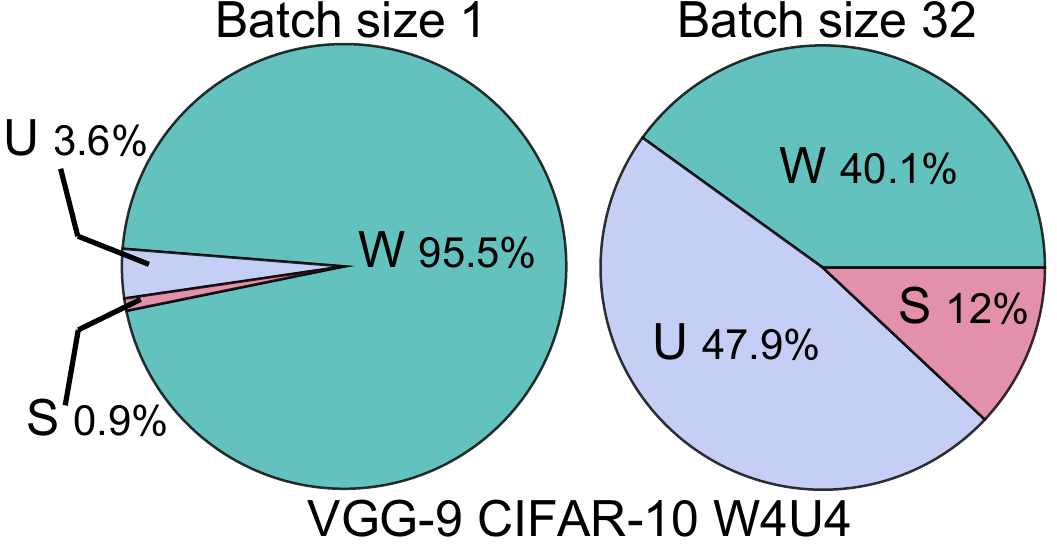} 
\\\vspace{-1mm}
{ (a)} & { (b)}
\vspace{-3mm}
\end{tabular}
\end{center}
  \caption{(a) Tradeoffs between accuracy (\textcolor{Dandelion}{yellow line}) and normalized computation energy across timesteps. (b) Memory energy breakdown: U for membrane potentials, W for weights, and S for output spikes.}
  \vspace{-1mm}
  \label{fig:ablation_2}
\end{figure}

In Fig.~\ref{fig:ablation_2}(a), we demonstrate that our method is effective with different timesteps using the \texttt{W2U2} VGG-9 model on CIFAR-10. While increasing timesteps slightly improves accuracy, the computation energy linearly increases with the timesteps. Lastly, we present a breakdown of memory energy in Fig.\ref{fig:ablation_2}(b). The results align with previous observations on membrane potential quantization. While the data movement energy of weights dominates at a batch size of 1, the memory energy cost of the membrane potential rises to 47.9\% of the total memory energy at a batch size of 32, underscoring the importance of membrane potential quantization.

\section{Conclusion}
In this paper, we presented the multiplier-less integer-based (MINT) quantization method for SNNs, which compresses both weights and membrane potentials. By sharing the quantization scale and transforming the LIF update equations, MINT significantly reduces the memory footprint of SNNs without incurring hardware overheads associated with scaling factors in conventional quantization. Our approach achieves accuracy comparable to full-precision baselines and other state-of-the-art SNN quantization methods. In conclusion, MINT provides a practical solution for minimizing memory footprint and hardware requirements of SNNs without compromising accuracy, opening up new possibilities for efficient SNN-based edge computing.
\vspace{-4mm}

\section*{\sc Acknowledgments}
\small This work was supported in part by CoCoSys, a JUMP2.0 center sponsored by DARPA and SRC, the National Science Foundation (CAREER Award, Grant \#2312366, Grant \#2318152), TII (Abu Dhabi), and the DoE MMICC center SEA-CROGS (Award \#DE-SC0023198).

\bibliographystyle{IEEEtran}
\bibliography{main}

\begin{thebibliography}{10}
\providecommand{\url}[1]{#1}
\csname url@samestyle\endcsname
\providecommand{\newblock}{\relax}
\providecommand{\bibinfo}[2]{#2}
\providecommand{\BIBentrySTDinterwordspacing}{\spaceskip=0pt\relax}
\providecommand{\BIBentryALTinterwordstretchfactor}{4}
\providecommand{\BIBentryALTinterwordspacing}{\spaceskip=\fontdimen2\font plus
\BIBentryALTinterwordstretchfactor\fontdimen3\font minus \fontdimen4\font\relax}
\providecommand{\BIBforeignlanguage}[2]{{%
\expandafter\ifx\csname l@#1\endcsname\relax
\typeout{** WARNING: IEEEtran.bst: No hyphenation pattern has been}%
\typeout{** loaded for the language `#1'. Using the pattern for}%
\typeout{** the default language instead.}%
\else
\language=\csname l@#1\endcsname
\fi
#2}}
\providecommand{\BIBdecl}{\relax}
\BIBdecl

\bibitem{spike_nature}
K.~Roy, A.~Jaiswal, and P.~Panda, ``Towards spike-based machine intelligence with neuromorphic computing,'' \emph{Nature}, 2019.

\bibitem{direct_snn}
Y.~Wu \emph{et~al.}, ``Direct training for spiking neural networks: Faster, larger, better,'' in \emph{AAAI}, 2019.

\bibitem{tssl}
W.~Zhang and P.~Li, ``Temporal spike sequence learning via backpropagation for deep spiking neural networks,'' \emph{NeurIPS}, 2020.

\bibitem{dtsnn}
Y.~Li, A.~Moitra, T.~Geller, and P.~Panda, ``Input-aware dynamic timestep spiking neural networks for efficient in-memory computing,'' \emph{DAC}, 2023.

\bibitem{imagenet}
J.~Deng, ``Imagenet: A large-scale hierarchical image database,'' \emph{CVPR}, 2009.

\bibitem{stq}
S.~S. Chowdhury, I.~Garg, and K.~Roy, ``Spatio-temporal pruning and quantization for low-latency spiking neural networks,'' in \emph{IJCNN}, 2021.

\bibitem{admm}
L.~Deng \emph{et~al.}, ``Comprehensive snn compression using admm optimization and activity regularization,'' \emph{TNNLS}, 2021.

\bibitem{stbp_q}
P.-Y. Tan and C.-W. Wu, ``A low-bitwidth integer-stbp algorithm for efficient training and inference of spiking neural networks,'' in \emph{ASPDAC}, 2023.

\bibitem{spinalflow}
S.~Narayanan, K.~Taht, R.~Balasubramonian, E.~Giacomin, and P.-E. Gaillardon, ``Spinalflow: An architecture and dataflow tailored for spiking neural networks,'' in \emph{ISCA}, 2020.

\bibitem{google_quant}
B.~Jacob \emph{et~al.}, ``Quantization and training of neural networks for efficient integer-arithmetic-only inference,'' in \emph{CVPR}, 2018.

\bibitem{whitepaper_quant}
R.~Krishnamoorthi, ``Quantizing deep convolutional networks for efficient inference: A whitepaper,'' \emph{arXiv:1806.08342}, 2018.

\bibitem{ptb}
J.-J. Lee, W.~Zhang, and P.~Li, ``Parallel time batching: Systolic-array acceleration of sparse spiking neural computation,'' in \emph{HPCA}, 2022.

\bibitem{hessianSnn}
H.~W. Lui and E.~Neftci, ``Hessian aware quantization of spiking neural networks,'' in \emph{ICONS}, 2021.

\bibitem{intSnn}
C.~J. Schaefer and S.~Joshi, ``Quantizing spiking neural networks with integers,'' \emph{ICONS}, 2020.

\bibitem{stdp_q}
N.~Rathi, P.~Panda, and K.~Roy, ``Stdp-based pruning of connections and weight quantization in spiking neural networks for energy-efficient recognition,'' \emph{TCAD}, 2018.

\bibitem{stdp_q2}
S.~Hu \emph{et~al.}, ``Quantized stdp-based online-learning spiking neural network,'' \emph{Neural Computing and Applications}, 2021.

\bibitem{qspin}
R.~V.~W. Putra and M.~Shafique, ``Q-spinn: A framework for quantizing spiking neural networks,'' in \emph{IJCNN}, 2021.

\bibitem{conv_q1}
A.~R. Voelker, D.~Rasmussen, and C.~Eliasmith, ``A spike in performance: Training hybrid-spiking neural networks with quantized activation functions,'' \emph{arXiv:2002.03553}, 2020.

\bibitem{conv_q2}
C.~Li, L.~Ma, and S.~Furber, ``Quantization framework for fast spiking neural networks,'' \emph{Frontiers in Neuroscience}, 2022.

\bibitem{sata}
R.~Yin, A.~Moitra, A.~Bhattacharjee, Y.~Kim, and P.~Panda, ``Sata: Sparsity-aware training accelerator for spiking neural networks,'' \emph{TCAD}, 2022.

\bibitem{vgg}
K.~Simonyan and A.~Zisserman, ``Very deep convolutional networks for large-scale image recognition,'' \emph{arXiv:1409.1556}, 2014.

\bibitem{resnet}
K.~He, X.~Zhang, S.~Ren, and J.~Sun, ``Deep residual learning for image recognition,'' in \emph{CVPR}, 2016.

\bibitem{krizhevsky2009learning}
A.~Krizhevsky \emph{et~al.}, ``Learning multiple layers of features from tinyimages,'' 2009.

\bibitem{cifar10dvs}
H.~Li, H.~Liu, X.~Ji, G.~Li, and L.~Shi, ``Cifar10-dvs: an event-stream dataset for object classification,'' \emph{Frontiers in neuroscience}, 2017.

\bibitem{cacti}
N.~Muralimanohar, R.~Balasubramonian, and N.~P. Jouppi, ``Cacti 6.0: A tool to model large caches,'' \emph{HP laboratories}, 2009.

\bibitem{bntt}
Y.~Kim and P.~Panda, ``Revisiting batch normalization for training low-latency deep spiking neural networks from scratch,'' \emph{Frontiers in neuroscience}, 2021.

\bibitem{spike-bp}
C.~Lee \emph{et~al.}, ``Enabling spike-based backpropagation for training deep neural network architectures,'' \emph{Frontiers in neuroscience}, 2020.

\end{thebibliography}

\end{document}